\pdfoutput=1

\documentclass[11pt]{article}

\usepackage[final]{acl}

\usepackage{microtype}
\usepackage{graphicx}
\usepackage{booktabs} 
\usepackage{caption}

\usepackage{amsmath}
\usepackage{amssymb}
\usepackage{mathtools}
\usepackage{amsthm}
\usepackage{times}
\usepackage{latexsym}
\usepackage{makecell}
\usepackage[most]{tcolorbox}
\usepackage{dblfloatfix}
\usepackage{enumitem}
\usepackage{subcaption}
\usepackage{CJKutf8}
\usepackage{lipsum}

\usepackage{array}
\usepackage{graphicx}
\usepackage{float}
\usepackage{xcolor}
\usepackage{booktabs}
\usepackage{multirow}
\usepackage{amsfonts}
\usepackage{amsmath}
\usepackage{setspace}
\usepackage{tabularx}
\usepackage[capitalize,noabbrev]{cleveref}

\theoremstyle{plain}

\theoremstyle{definition}

\theoremstyle{remark}

\usepackage[textsize=tiny]{todonotes}

\title{LangMark: A Multilingual Dataset for Automatic Post-Editing}

\author{%
  Diego Velazquez$^{1}$\quad
  Mikaela Grace$^{1}$\quad
  Konstantinos Karageorgos$^{1}$\quad
  Lawrence Carin$^{2}$\\
  \textbf{Aaron Schliem}$^{1}$\quad
  \textbf{Dimitrios Zaikis}$^{1}$\quad
  \textbf{Roger Wechsler}$^{1}$\quad \\ \\
  {
    $^1$Welocalize\hspace{1em}
    $^2$Duke University
  }
}

\begin{document}
\maketitle




\begin{abstract}
Automatic post-editing (APE) aims to correct errors in machine-translated text, enhancing translation quality, while reducing the need for human intervention. Despite advances in neural machine translation (NMT), the development of effective APE systems has been hindered by the lack of large-scale multilingual datasets specifically tailored to NMT outputs. To address this gap, we present and release  \textbf{LangMark}\footnote{https://zenodo.org/records/15553365}, a new human-annotated multilingual APE dataset for English translation to seven languages: Brazilian Portuguese, French, German, Italian, Japanese, Russian, and Spanish. The dataset has 206,983 triplets, with each triplet consisting of a source segment, its NMT output, and a human post-edited translation. Annotated by expert human linguists, our dataset offers both language diversity and scale.
Leveraging this dataset, we empirically show that \emph{Large Language Models} (LLMs) with few-shot prompting can effectively perform APE, improving upon leading commercial and even proprietary machine translation systems. We believe that this new resource will facilitate the future development and evaluation of APE systems.
\end{abstract}

\section{Introduction}
Machine translation has become increasingly efficient and effective thanks to the development of ever-larger transformer models \citep{vaswani2017attention}. Recent advances in \emph{Large Language Models} (LLMs) have significantly influenced the field, enabling more fluent and contextually accurate translations~\citep{zhu-etal-2024-multilingual, zhang-etal-2023-machine, li-etal-2024-towards-demonstration, briakou2024translating}. Studies have shown that LLMs can match or even outperform specialized systems in various Natural Language Processing (NLP) tasks~\citep{radford2019language, touvron2023llama, wang2022language}.

\begin{figure}[t]
    \centering
    \includegraphics[width=0.48\textwidth]{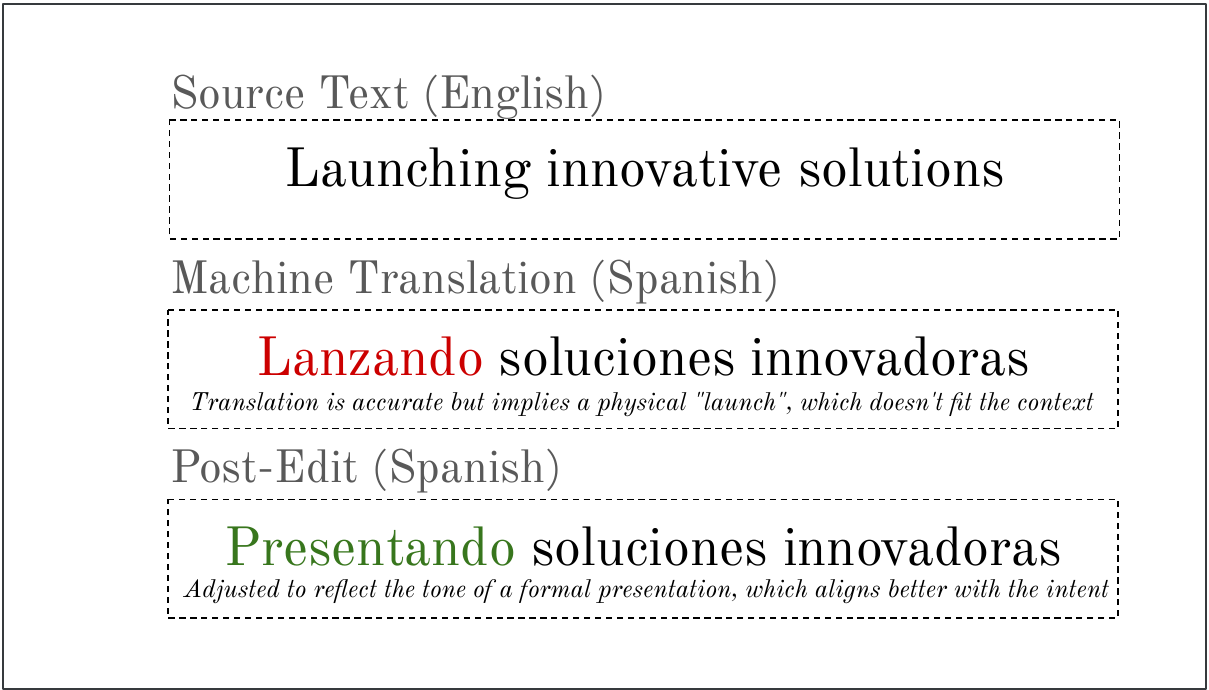}
    \caption{Example of a triplet in an automatic post-editing task.}
    \label{fig:post-edit-example}
\end{figure}

Despite these advances, machine-translated text often still contains errors that require correction to meet the quality standards expected in professional translations. Automatic Post-Editing (APE) aims to automatically correct these errors in MT output, improving translation quality while reducing the need for human intervention \citep{knight1994automated}. Modern APE models take the source text and machine-translated text as input and produce the post-edited text with the necessary changes as output. We refer to these components as triplets: \emph{source}, \emph{translated}, and \emph{post-edited} segments (see Figure \ref{fig:post-edit-example}).

Recently, automatic post-editing has shown success on Statistical Machine Translation (SMT) outputs \citep{junczys-dowmunt-grundkiewicz-2018-ms, correia-martins-2019-simple}, but even strong APE models face significant challenges on modern NMT outputs \citep{chatterjee-etal-2019-findings, chatterjee-etal-2018-findings, ive-etal-2020-post}. For instance, \citet{chollampatt2020pedra} demonstrated that fine-tuned Transformer models can improve upon state-of-the-art NMT, yet their SubEdits dataset (161K triplets) is limited to a single language pair (English-German). This highlights the need for larger, multilingual datasets to advance APE research on NMT outputs.

In an effort to address this gap, we introduce \textbf{LangMark}; a new multilingual, human-post-edited APE dataset comprising 206,983 triplets from English to seven languages:  Brazilian Portuguese (BR), French (FR), German (DE), Italian (IT), Japanese (JP), Russian (RU), and Spanish (ES) (see Table \ref{tab:dataset-statistics}). Each triplet consists of a source segment in English, its NMT output, and a human post-edited translation. Labeled by expert linguists, this dataset offers both language diversity and scale, making it, to the best of our knowledge, the largest human-post-edited dataset for APE on NMT outputs.

\begin{table}[t]
\centering
\caption{Number of triplets and average \emph{source}, \emph{NMT} and \emph{Post Edited} tokens (tokenized using \textit{tiktoken}\protect\footnotemark
) per triplet for all languages in LangMark.}
\resizebox{\columnwidth}{!}{
\begin{tabular}{lrr|r|rrr}
\toprule
\textbf{Locale} & \textbf{Triplets} & \multicolumn{3}{c}{\textbf{Tokens Per Triplet (AVG)}} \\ 
\cmidrule{3-5}
                &                   & \textbf{Source} & \textbf{NMT} & \textbf{PE} \\ 
\midrule
EN-DE           & 33,308           & 16.12          & 21.73        & 21.72       \\ 
EN-ES         & 32,799           & 16.58          & 20.80        & 21.16       \\ 
EN-FR           & 33,027           & 16.38          & 22.16        & 22.35       \\ 
EN-IT           & 32,512           & 16.42          & 23.47        & 23.71       \\ 
EN-JP           & 28,170           & 15.26          & 26.34        & 27.30       \\ 
EN-BR           & 31,981           & 16.52          & 20.36        & 20.30       \\ 
EN-RU           & 8,648            & 14.90          & 20.40        & 21.23       \\ 
\bottomrule
\end{tabular}
}
\label{tab:dataset-statistics}
\end{table}

\footnotetext{\url{https://github.com/openai/tiktoken}}

Leveraging this dataset, we empirically show that LLMs with few-shot prompting can effectively perform APE, improving upon leading commercial and proprietary MT systems. Our experiments highlight the potential of combining large-scale, high-quality datasets with advanced LLMs to enhance translation quality across multiple languages. Moreover, this work examines a critical aspect of APE: the model’s capability to discern whether a segment requires editing, which has not always been explicitly addressed in prior research.

The contributions of this work can be summarized as follows:

\begin{enumerate}
    \item We present and release \textbf{LangMark}, a new, human-annotated, multilingual dataset with over 200,000 triplets across seven languages, that serves as a strong benchmark for APE tasks.
    \item Leveraging this dataset, we show that LLMs with few-shot prompting can effectively perform APE to improve upon NMT outputs even from proprietary MT systems.
    \item We provide a comprehensive analysis of the dataset and the performance of LLMs on APE tasks, offering insights for future research.
\end{enumerate}

\section{Related Work}

This section reviews previous research on automatic post-editing, focusing on recent advancements involving Large Language Models. We also examine retrieval methods for few-shot in-context learning and discuss relevant datasets used for post-editing tasks.

\subsection{Automatic Post-Editing}

Automatic post-editing aims to automatically correct errors in machine-translated text, improving translation quality without human intervention. A great amount of prior research has focused on the development of neural models for the APE task \citep{vu2018automatic, shterionov2020roadmap, chatterjee2019automatic, gois2020learning, correia-martins-2019-simple, voita-etal-2019-context, chollampatt-etal-2020-automatic, do2021review}. \citet{shterionov2020roadmap} presented a comprehensive roadmap for APE, highlighting challenges and potential directions for future research. \citet{chatterjee2019automatic} explored the use of deep learning techniques for APE while \citet{gois2020learning} investigated the use of automatic ordering techniques to refine translations. \citet{correia-martins-2019-simple} proposed a simple yet effective neural model for APE using transfer learning, demonstrating promising results.

\citet{voita-etal-2019-context} introduced a context-aware approach to APE, incorporating source context information into the neural model to generate more accurate post-edits. \citet{chollampatt-etal-2020-automatic} examined the use of transformer-based models for APE to improve overall translation quality for NMT models, investigating the effects of various factors in the APE task. \citet{do2021review} provided an overview of various techniques and approaches in the field of APE, covering both traditional and neural-based methods. Overall, these studies (and many references therein) have explored different architectures, learning strategies, and contextual information integration in neural models to improve the quality of post-edited translations.

\begin{table}[b!]
    \centering
    \resizebox{\columnwidth}{!}{
    \begin{tabular}{lcll}
        \toprule
        \textbf{Dataset} & \textbf{Lang.} & \textbf{Size} & \textbf{Domain} \\ 
        \midrule
        \makecell[l]{WMT'18 APE \\ \citep{chatterjee-etal-2018-findings}} & EN-DE & 15K & IT \\ 
        \makecell[l]{WMT'19 APE \\ \citep{chatterjee-etal-2019-findings}} & EN-RU & 17K & IT \\ 
        \makecell[l]{WMT'23 APE \\ \citep{bhattacharyya2023findings}} & EN-MR & 18K & Mixed \\ 
        \midrule
        \makecell[l]{QT21 \\ \citep{specia-etal-2017-translation}} & EN-LV & 21K & Life Sciences \\ 
        \midrule
        \makecell[l]{APE-QUEST \\ \citep{ive-etal-2020-post}} & \begin{tabular}[c]{@{}c@{}}EN-NL \\ EN-FR \\ EN-PT\end{tabular} & 
        \begin{tabular}[c]{@{}c@{}}11K \\ 10K \\ 10K\end{tabular} & Legal \\ 
        \midrule
        \makecell[l]{SubEdits \\ \citep{chollampatt2020pedra}} & EN-DE & 161K & Subtitles \\ 
        \midrule
        \makecell[l]{eSCAPE (Artificial) \\ \citep{negri2018online}} & \begin{tabular}[c]{@{}c@{}}EN-DE \\ EN-IT \\ EN-RU\end{tabular} & 
        \begin{tabular}[c]{@{}c@{}}7.2M \\ 3.3M \\ 7.7M\end{tabular} & Mixed \\ 
        \midrule
        \makecell[l]{LangMark\\ (this work)} & 
        \begin{tabular}[c]{@{}c@{}}EN-DE \\ EN-ES \\ EN-FR \\ EN-IT \\ EN-JP \\ EN-BR \\ EN-RU\end{tabular} & 
        \begin{tabular}[c]{@{}c@{}}33.3K \\ 32.7K \\ 33.1K \\ 32.5K \\ 28.1K \\ 31.9K \\ 8.6K\end{tabular} & 
        Marketing \\
        \bottomrule
    \end{tabular}
    }
    \caption{Datasets for automatic post-editing on NMT outputs. All but eSCAPE offer human labels.}
    \label{tab:datasets-ape}
\end{table}

\subsection{Leveraging Large Language Models for Post-Editing}

There has been growing interest in leveraging LLMs for post-editing. For example, \citet{vidal-etal-2022-automatic} explored the use of GPT-3 for post-editing using glossaries, while \citet{raunak2023leveraging} investigated the use of GPT-4 for automatic post-editing of neural machine translation outputs. Their work focuses on rectifying errors in NMT outputs without preliminary quality assessment, aiming to enhance translation quality directly.

\citet{ki2024guiding} further enhances machine translation by guiding large language models to post-edit MT outputs using fine-grained feedback from error annotations. Their experiments across multiple language pairs demonstrate that both zero-shot prompted and fine-tuned LLMs benefit from this approach, effectively addressing specific translation errors and improving translation metrics.

\textbf{In parallel}, \citet{treviso2024xtower} propose using quality estimation (QE) thresholds to decide whether the original MT output even needs editing, combining LLM-based correction with a preliminary QE-based decision step. \textbf{Additionally}, \citet{koneru-etal-2024-contextual} and \citet{li-etal-2025-enhancing-large} demonstrate that incorporating \emph{document-level} context can further refine LLM-driven APE, yielding higher translation quality than sentence-level post-editing alone.

While these works make significant contributions to the exploration of LLMs for post-editing, they do not constitute a benchmark for evaluating the multilingual post-editing capabilities of LLMs. In contrast, we believe that \textbf{LangMark}, coupled with the experiments presented in this paper, can serve as a robust benchmark for this purpose, enabling a more comprehensive assessment of LLM performance across multiple languages.

\begin{table*}[!t]
\centering
\caption{Machine translation performance across languages for different NMT engines on all triplets of the \textbf{LangMark} dataset.}
\label{tab:mt_system_comparison}
\resizebox{\textwidth}{!}{
\begin{tabular}{l|cc|cc|cc|cc|cc|cc|cc}
\toprule
\textbf{MT Engine} & \multicolumn{2}{c|}{\textbf{EN-DE}} & \multicolumn{2}{c|}{\textbf{EN-ES}} & \multicolumn{2}{c|}{\textbf{EN-FR}} & \multicolumn{2}{c|}{\textbf{EN-IT}} & \multicolumn{2}{c|}{\textbf{EN-JP}} & \multicolumn{2}{c|}{\textbf{EN-PT}} & \multicolumn{2}{c}{\textbf{EN-RU}} \\
\midrule
\textbf{Metric} & CHRF & TER$\downarrow$ & CHRF & TER$\downarrow$ & CHRF & TER$\downarrow$ & CHRF & TER$\downarrow$ & CHRF & TER$\downarrow$ & CHRF & TER$\downarrow$ & CHRF & TER$\downarrow$ \\
\midrule
Google Translate     & 73.95 & 42.16 & 79.79 & 27.54 & 76.57 & 33.14 & 79.80 & 28.98 & 62.11 & 78.64 & 83.70 & 21.12 & 64.34 & 53.46 \\
DeepL      & 73.03 & 43.15 & 75.01 & 33.70 & 74.74 & 36.27 & 76.96 & 33.05 & 55.26 & 91.52 & 83.93 & 22.68 & 67.74 & 47.41 \\
Microsoft Translator         & 75.74 & 40.35 & 80.32 & 27.55 & 76.07 & 34.29 & 82.57 & 25.29 & 62.82 & 84.06 & 84.97 & 20.35 & 64.38 & 54.39 \\
Amazon Translate    & 73.70 & 43.13 & 79.01 & 29.78 & 76.27 & 34.42 & 81.66 & 26.52 & 60.93 & 86.62 & 84.27 & 21.96 & 62.65 & 56.00 \\
\midrule
Proprietary MT (this dataset) & \textbf{81.09} & \textbf{31.35} & \textbf{86.04} & \textbf{19.39} & \textbf{81.54} & \textbf{26.99} & \textbf{89.73} & \textbf{14.58} & \textbf{69.77} & \textbf{74.66} & \textbf{89.13} & \textbf{14.64} & \textbf{68.45} & \textbf{45.54} \\
\bottomrule
\end{tabular}
}
\end{table*}

\subsection{Datasets for Automatic Post-Editing}

Several earlier works focused on post-editing for statistical machine translation (SMT). The largest collection of human post-edits on SMT outputs was released by \citet{zhechev2012machine}, comprising 30,000 to 410,000 triplets across 12 language pairs. While SMT-based APE often showed impressive gains \citep{junczys-dowmunt-2017-amu, tebbifakhr-etal-2018-multi}, transitioning to NMT introduced new challenges, and some studies found only marginal improvements \citep{chatterjee-etal-2019-findings, junczys-dowmunt-grundkiewicz-2018-ms}.

To support NMT-based APE, researchers have turned to synthetic data generation \citep{junczys-dowmunt-grundkiewicz-2016-log, freitag-etal-2019-ape, specia-etal-2017-translation, negri2018online, li-etal-2024-towards-demonstration}. However, purely artificial datasets sometimes fail to capture the nuanced edits required by advanced NMT systems. Human-labeled data are thus crucial, yet existing resources—such as the WMT APE shared tasks \citep{chatterjee-etal-2018-findings, chatterjee-etal-2019-findings} or SubEdits \citep{chollampatt2020pedra}—tend to be either limited in scale or language diversity. Table~\ref{tab:datasets-ape} summarizes these datasets

These datasets contribute valuable resources for studying post-editing but are limited in language diversity or scale when providing human annotations. In contrast, the dataset featured in this work is a multilingual, human-annotated corpus consisting of translations from English to seven languages, with over  200,000 triplets. To the best of our knowledge, \textbf{LangMark} is the largest multilingual, human-annotated dataset for APE on NMT outputs.

\section{LangMark Dataset}
The absence of large-scale, multilingual, human-annotated corpora for post-editing NMT outputs presents a gap in the resources available for advancing APE research. To address this limitation, we introduce \textbf{LangMark}, a new dataset comprising over 200,000 triplets across seven language pairs: English to Japanese (JP), Russian (RU), Brazilian Portuguese (BR), Spanish (ES), French (FR), Italian (IT), and German (DE).

\begin{figure}[t!]
    \centering
    \includegraphics[width=0.48\textwidth]{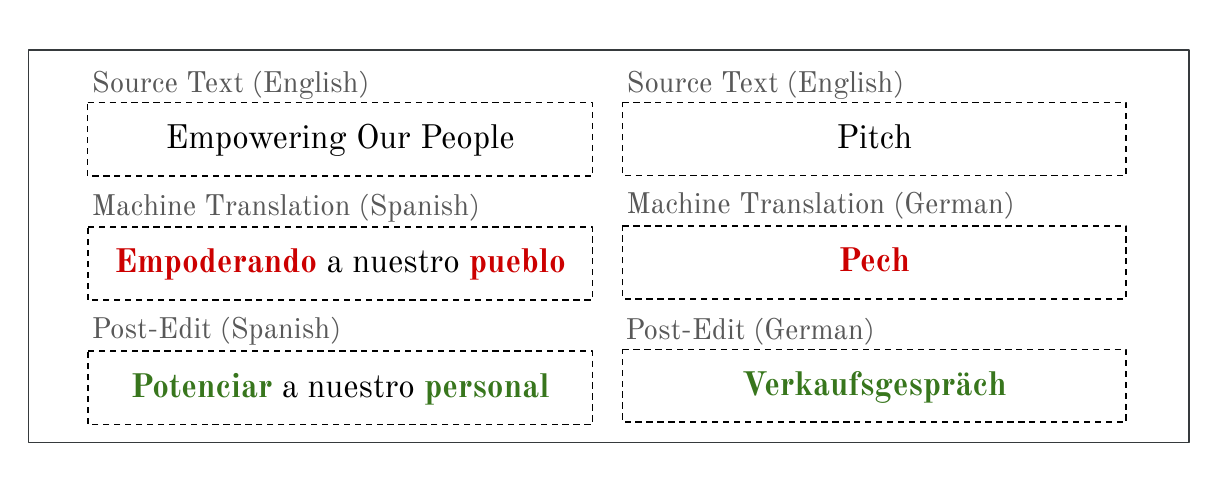}
    \caption{Two triplets from the \textbf{LangMark} dataset. These examples illustrate the nuanced nature of the required corrections. While the translations provided by the NMT engine are not inherently incorrect, they are inappropriate given the context of the source material (official marketing documents). For example, ``our people'' was misinterpreted as ``our nation/community'' in Spanish, and ``pitch'' was translated based on the meaning of ``tar'' in German instead of its intended meaning in a business context.}
    \label{fig:dataset-sample}
\end{figure}

The \textbf{LangMark} dataset contains a large number of segments that require models to make nuanced edits, which makes it challenging as a benchmark. NMT outputs in the dataset are often technically correct but fail to align with the intended context (see \cref{fig:dataset-sample}). To successfully post-edit these samples the model has to demonstrate contextual understanding. You can find some examples of post-edited segments in \ref{sec:pe_examples}.

\subsection{Dataset Source}
\label{sec:dataset-source}
The \textbf{LangMark} dataset is sourced from various Smartsheet\footnote{\url{https://www.smartsheet.com}} documents, a platform designed for collaborative work management. These documents, which are marketing-related, were first segmented by a translation management system (TMS) into intuitive units (often sentences or short phrases) before translation. This standard industry practice ensures efficient processing, storage, and translation workflows. The triplets were then randomly selected from 967 unique files.

To protect sensitive information, we used Google's dlp\footnote{\url{https://cloud.google.com/dlp}} tool, specifically designed to identify and remove personally identifiable information (PII) and other sensitive data. We also removed duplicate triplets for each language pair; apart from this preprocessing step, the segments are presented in their original form, reflecting the nature of real-world industry data. We consider this characteristic a positive feature, as it allows the evaluation of model performance on authentic, unaltered data, closely mirroring practical use cases in the industry.

\subsection{Neural Machine Translation}
The dataset features NMT outputs generated by a proprietary MT system tailored to Smartsheet, along with post-edited translations produced by expert linguists. Because these proprietary machine translation engines are trained on in-domain data, they can be particularly strong in narrow areas, providing high-quality outputs that set a rigorous baseline. This ensures that automatic post-editing (APE) systems are evaluated against a robust benchmark, making any improvements reflective of real-world challenges. Table \ref{tab:mt_system_comparison} shows the difference in performance between the NMT comprised in \textbf{LangMark} and commercial MT systems.

\begin{figure*}[t!]
    \centering
    \begin{subfigure}[b]{0.32\textwidth}
        \centering
        \includegraphics[width=\textwidth]{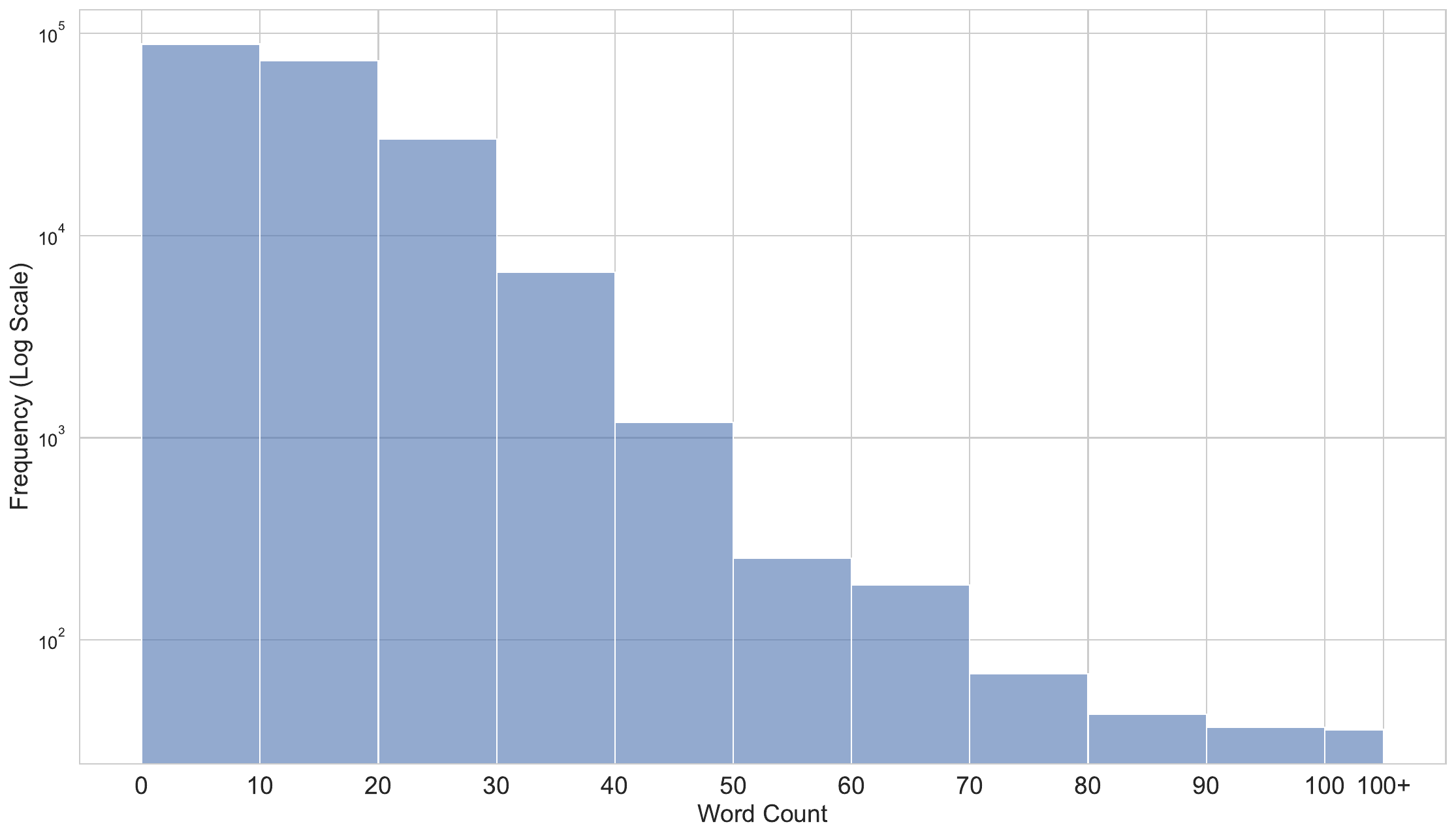}
        \caption{Source segment word counts.}
        \label{fig:word-count}
    \end{subfigure}
    \hfill
    \begin{subfigure}[b]{0.32\textwidth}
        \centering
        \includegraphics[width=\textwidth]{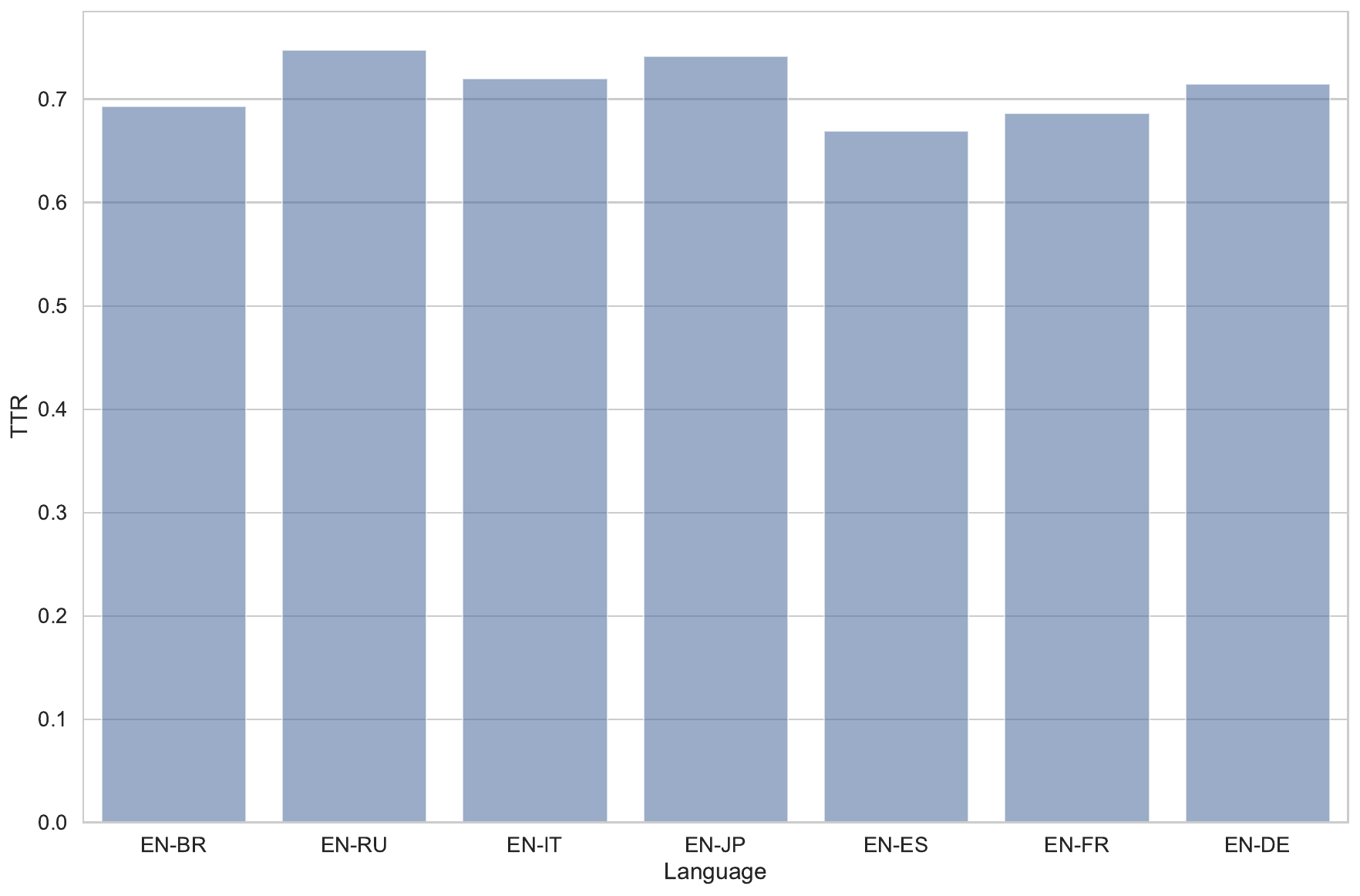}
        \caption{Token/Type Ratio per language.}
        \label{fig:ttr}
    \end{subfigure}
    \hfill
    \begin{subfigure}[b]{0.32\textwidth}
        \centering
        \includegraphics[width=\textwidth]{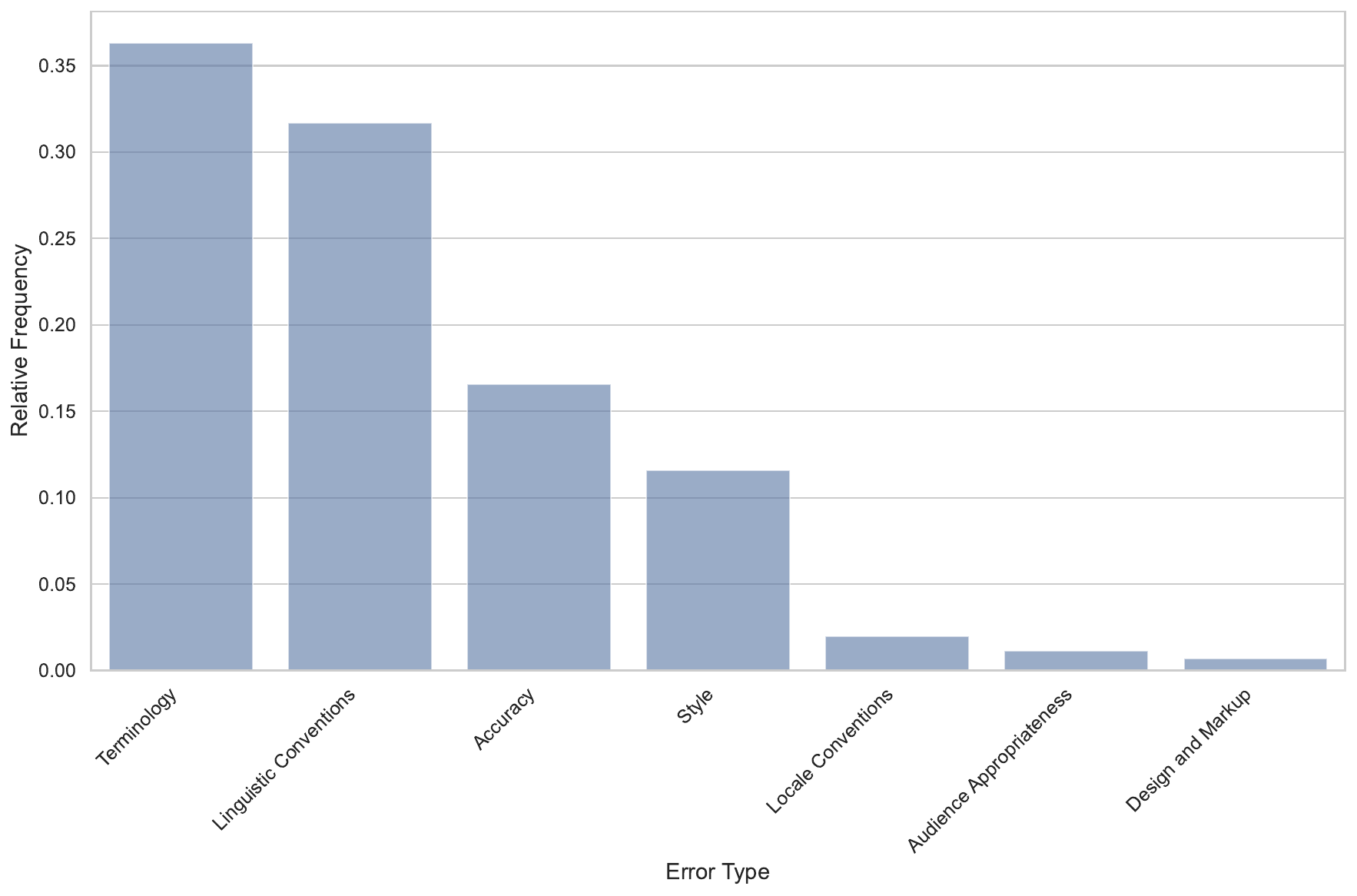}
        \caption{Relative frequency of MQM errors.}
        \label{fig:error-frequency}
    \end{subfigure}
    \caption{Dataset statistics: (a) distribution of word counts for source segments, (b) lexical diversity measured using window-based TTR across languages, and (c) relative frequency of MQM error types in the pre-translations that need correction.}
    \label{fig:dataset-stats}
\end{figure*}

\subsection{Dataset Statistics}
The dataset comprises 206,983 triplets from English to seven languages, with each triplet containing a source segment, its NMT output, and a human post-edited translation.

Figure~\ref{fig:dataset-stats} presents key dataset statistics, including segment length distribution, lexical diversity across languages, and the distribution of MQM error types\footnote{The error type assignment was done using an internal tool.}, highlighting the dataset's balanced composition, linguistic variability and error type diversity.

\subsection{Linguist Qualifications}
\label{sec:linguists-qualifications}

We source and deploy linguists with credentials such as degrees in linguistics or translation, native-level fluency in the target language, and strong cultural knowledge—preferably as in-country professionals. All linguists are required to have over five years of industry experience, advanced proficiency in translation tools, and a proactive approach to continuous improvement. Additionally, they must specialize in translating and post-editing content within specific subject matter domains, often with more than three years of expertise in these areas. Following onboarding, linguists receive ongoing support and training to maintain quality, monitored through structured Language Quality Assessments (LQAs). Based on these evaluations, further training or reassignment ensures alignment with project needs. For information on linguist compensations and instructions, see \ref{sec:comp} and \ref{sec:instructions} respectively.

\subsection{Post-Editing Process}
In constructing the dataset, our human post-editors (see \cref{sec:linguists-qualifications}), refined the raw NMT output within a Translation Management System (TMS). They made the necessary edits to ensure accuracy, adherence to stylistic and terminology standards, and overall readability, rather than rewriting the translation. The editors have access to glossaries, do-not-translate lists, and any necessary domain-specific materials. Common corrections addressed capitalization, punctuation, spacing, omissions, word order, morphological agreement, locale conventions, and terminology consistency. This process ensures that the final post-edited translations are aligned with client and domain expectations.

\section{Experimental Setup}
To evaluate the performance of the models, we split the dataset into ``training'' and testing sets, with 90\% of the triplets used as potential examples to be retrieved and the remaining 10\% reserved for experiments. The split is performed randomly for each language pair, ensuring a proportional representation of all languages.

We adopt this split and retrieval approach because even top-performing LLMs struggle to surpass the proprietary neural machine translation (NMT) engines in this dataset when presented with no context. The nuanced nature of the required edits makes zero-shot approaches insufficient, which motivates the inclusion of in-context examples to guide the model’s post-editing decisions. Furthermore, by limiting results to the test set, we make benchmarking on this dataset more affordable for future users. We evaluate all models with 20-shot prompts. For completeness, zero-shot results are provided in the Appendix~\ref{sec:zero-shot}.

\subsection{Retrieval}
We constructed the retrieval database by embedding the source segments using OpenAI’s ``text-embedding-3-small'' model.\footnote{\url{https://platform.openai.com/docs/models/}} Each source segment is stored alongside its corresponding post-edited translation. For retrieval during experiments, the source segment to be post-edited is embedded, and cosine similarity is used to identify the twenty most similar source-human post-edit pairs from the database. Retrieval is conducted within the same language pair, ensuring that no cross-lingual retrieval occurs.

\begin{figure}[H]
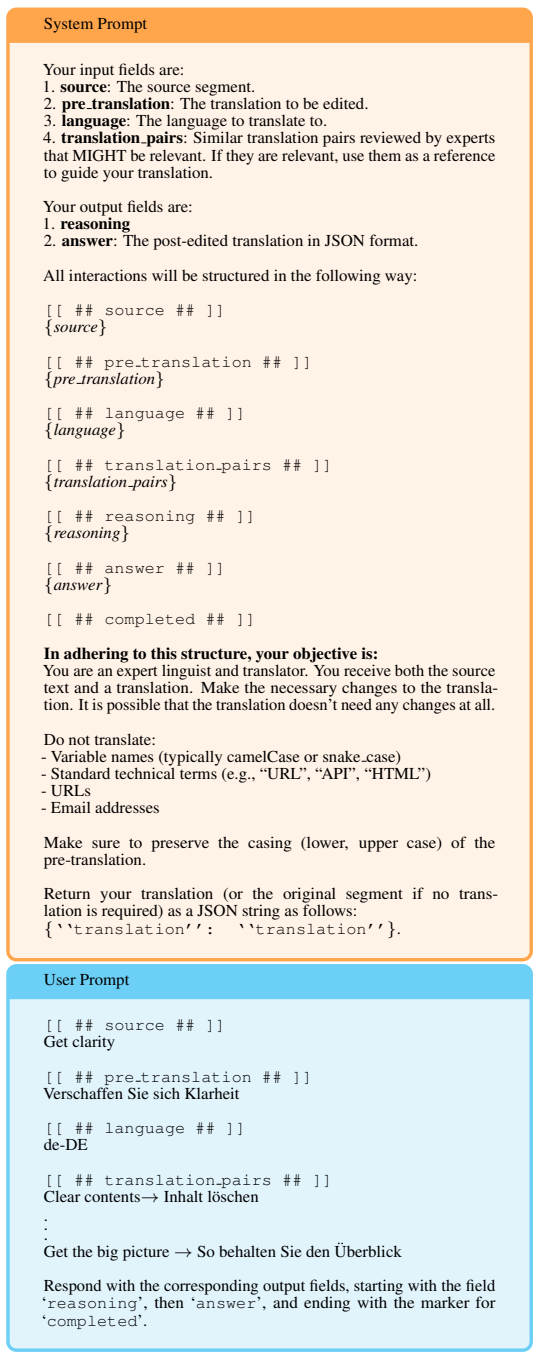

\centering
\scriptsize
 \scalebox{0.9}{
\begin{tcolorbox}[colback=orange!10, colframe=orange!70, coltitle=black,title=System Prompt]

\setstretch{0.9} 
Your input fields are: \\
1. \textbf{source}: The source segment. \\
2. \textbf{pre\_translation}: The translation to be edited. \\
3. \textbf{language}: The language to translate to. \\
4. \textbf{translation\_pairs}: Similar translation pairs reviewed by experts that MIGHT be relevant. If they are relevant, use them as a reference to guide your translation. \\

Your output fields are: \\
1. \textbf{reasoning} \\
2. \textbf{answer}: The post-edited translation in JSON format. \\

All interactions will be structured in the following way: \\

\texttt{[[ \#\# source \#\# ]]} \\ \{\textit{source}\} \\ \\
\texttt{[[ \#\# pre\_translation \#\# ]]} \\ \{\textit{pre\_translation}\} \\ \\
\texttt{[[ \#\# language \#\# ]]} \\ \{\textit{language}\} \\ \\
\texttt{[[ \#\# translation\_pairs \#\# ]]} \\ \{\textit{translation\_pairs}\} \\ \\
\texttt{[[ \#\# reasoning \#\# ]]} \\ \{\textit{reasoning}\} \\ \\
\texttt{[[ \#\# answer \#\# ]]} \\ \{\textit{answer}\} \\ \\ 
\texttt{[[ \#\# completed \#\# ]]} \\

\textbf{In adhering to this structure, your objective is:} \\
You are an expert linguist and translator. You receive both the source text and a translation. Make the necessary changes to the translation. It is possible that the translation doesn't need any changes at all. \\
\\
Do not translate: \\
\quad - Variable names (typically camelCase or snake\_case) \\
\quad - Standard technical terms (e.g., ``URL'', ``API'', ``HTML'') \\
\quad - URLs \\
\quad - Email addresses \\
\\
Make sure to preserve the casing (lower, upper case) of the pre-translation. \\
\\
Return your translation (or the original segment if no translation is required) as a JSON string as follows: \\
\texttt{\{``translation'': ``translation''\}}.
\end{tcolorbox}
}
 \scalebox{0.9}{
\begin{tcolorbox}[colback=cyan!10, colframe=cyan!50, coltitle=black,title=User Prompt]
\setstretch{0.9} 
\texttt{[[ \#\# source  \#\# ]]} \\ 
Get clarity \\ \\
\texttt{[[ \#\# pre\_translation \#\# ]]} \\ Verschaffen Sie sich Klarheit \\ \\
\texttt{[[ \#\# language \#\# ]]} \\ de-DE \\ \\ 
\texttt{[[ \#\# translation\_pairs \#\# ]]} \\
\quad Clear contents$\rightarrow$ Inhalt löschen \\
\quad \vdots \\
\quad Get the big picture $\rightarrow$ So behalten Sie den Überblick \\

Respond with the corresponding output fields, starting with the field `\texttt{reasoning}', then `\texttt{answer}', and ending with the marker for `\texttt{completed}'.
\end{tcolorbox}
}
\caption{Structure of the few-shot prompting format used for LLMs. If the model's API does not support a system prompt we simply prepend it to the user prompt.}
\label{fig:prompt-format}
\end{figure}

\subsection{Models and Prompting}
We evaluate the performance of both open-source and closed-source models in our experiments. To facilitate this, we leverage the \texttt{dspy} library \citep{khattab2024dspy, khattab2022demonstrate}, which integrates with \texttt{LiteLLM}\footnote{\url{https://www.litellm.ai/}} to manage API requests to the various models. For open-source models, we utilize HuggingFace endpoints\footnote{\url{https://endpoints.huggingface.co/}} to set up and manage the necessary infrastructure to process requests.

All models are evaluated using the same 20-shot prompting setup. Specifically, for each segment to be post-edited, we include 20 pairs of source segments and their human post-edited version in the prompt. This ensures a uniform evaluation framework across all models. The prompt format used in our experiments is illustrated in \cref{fig:prompt-format}.

\begin{table*}[t!]
\scriptsize
\centering  
\caption{CHRF scores for different models and languages when performing APE on the test set. Scores are compared across models, with the proprietary MT serving as the baseline.}
\resizebox{0.8\textwidth}{!}{
\begin{tabular}{lccccccc}

\toprule
 & & \multicolumn{6}{c}{\textbf{Languages}} \\
\cmidrule(lr){2-8}
\textbf{Model} & \textbf{EN-RU} & \textbf{EN-BR} & \textbf{EN-JP} & \textbf{EN-IT} & \textbf{EN-FR} & \textbf{EN-ES} & \textbf{EN-DE} \\
\midrule
\textbf{Baseline} & 68.90 & \textbf{89.44} & 70.22 & 89.58 & 81.96 & 86.07 & 81.29 \\
\midrule
\textbf{Gemini-1.5 Flash} & 68.92 & 89.18 & 71.69 & 89.40 & 82.20 & 86.24 & 81.01 \\
\textbf{Gemini-1.5 Pro} & 67.73 & 87.65 & 68.92 & 85.68 & 80.46 & 85.01 & 77.88 \\
\textbf{Claude 3.5-Sonnet} & 68.63 & 86.47 & 67.14 & 85.10 & 80.31 & 82.73 & 78.44 \\
\textbf{Claude 3.5-Haiku} & 69.08 & 88.81 & 71.64 & 88.76 & 82.21 & 86.08 & 80.66 \\
\textbf{GPT-4o mini} & 68.55 & 87.73 & 68.47 & 87.47 & 81.45 & 84.94 & 79.81 \\
\textbf{GPT-4o} & 69.68 & 89.21 & \textbf{73.94} & \textbf{89.79} & \textbf{82.75} & \textbf{86.62} & \textbf{81.41} \\
\midrule
\multicolumn{8}{c}{\textbf{Open Source}} \\
\midrule
\textbf{Llama 3.1-70B} & 69.55 & 86.82 & 68.37 & 86.80 & 80.97 & 83.75 & 79.12 \\
\textbf{Qwen2.5-72B} & \textbf{70.13} & 89.03 & 72.93 & 89.10 & 82.34 & 86.44 & 81.16 \\
\bottomrule
\end{tabular}
}
\label{tab:chrf-scores}
\end{table*}

\section{Results and Discussion}
We benchmark the performance of various models on the \textbf{LangMark} test set and discuss broader challenges when evaluating performance on automatic post-editing (APE) tasks. While we have chosen CHRF \citep{popovic-2015-chrf} to show performance in the main text, we report other metrics in the Appendix (\ref{sec:more-metrics}).

\subsection{Model Performance} 
Table~\ref{tab:chrf-scores} presents the CHRF scores of various closed- and open-source models performing automatic post-editing on the \textbf{LangMark} test set using $n$-shot prompting ($n=20$). The results indicate that \textit{GPT-4o} consistently achieves the highest CHRF scores, being the only closed-source model that consistently improves the NMT output (except for Portuguese), especially in languages where more edits are required (i.e., Japanese and Russian). We also benchmark two open-source models of the \textit{Qwen} and \textit{Llama} family. We found that the performance of the \textit{Qwen} model is impressive for its size, rivaling the best closed-source models and even performing best in Russian. 

The strong performance of certain models should not overshadow the broader challenge presented by this dataset. Note that all of the models (except \textit{GPT-4o}) are unable to improve on the NMT baseline, which emphasizes the strength of this dataset as a benchmark for APE.

\subsection{To Edit or Not to Edit}
A critical aspect of automatic post-editing (APE) lies in determining when edits are necessary: some segments require changes while others are best left unchanged. This introduces a classification problem that the model must solve. As NMT systems continue to improve, the challenge shifts. High-performing NMT systems produce outputs that are closer to human translations. In this context, a language model that makes only a few highly accurate edits can achieve better evaluation scores than one that identifies more issues but fails to correct them in the exact manner a human would. This raises a crucial question for evaluating APE systems: \emph{``How conservative should models be when deciding that an edit is required?''}

\begin{figure}[b!]
    \centering
    \includegraphics[width=0.48\textwidth]{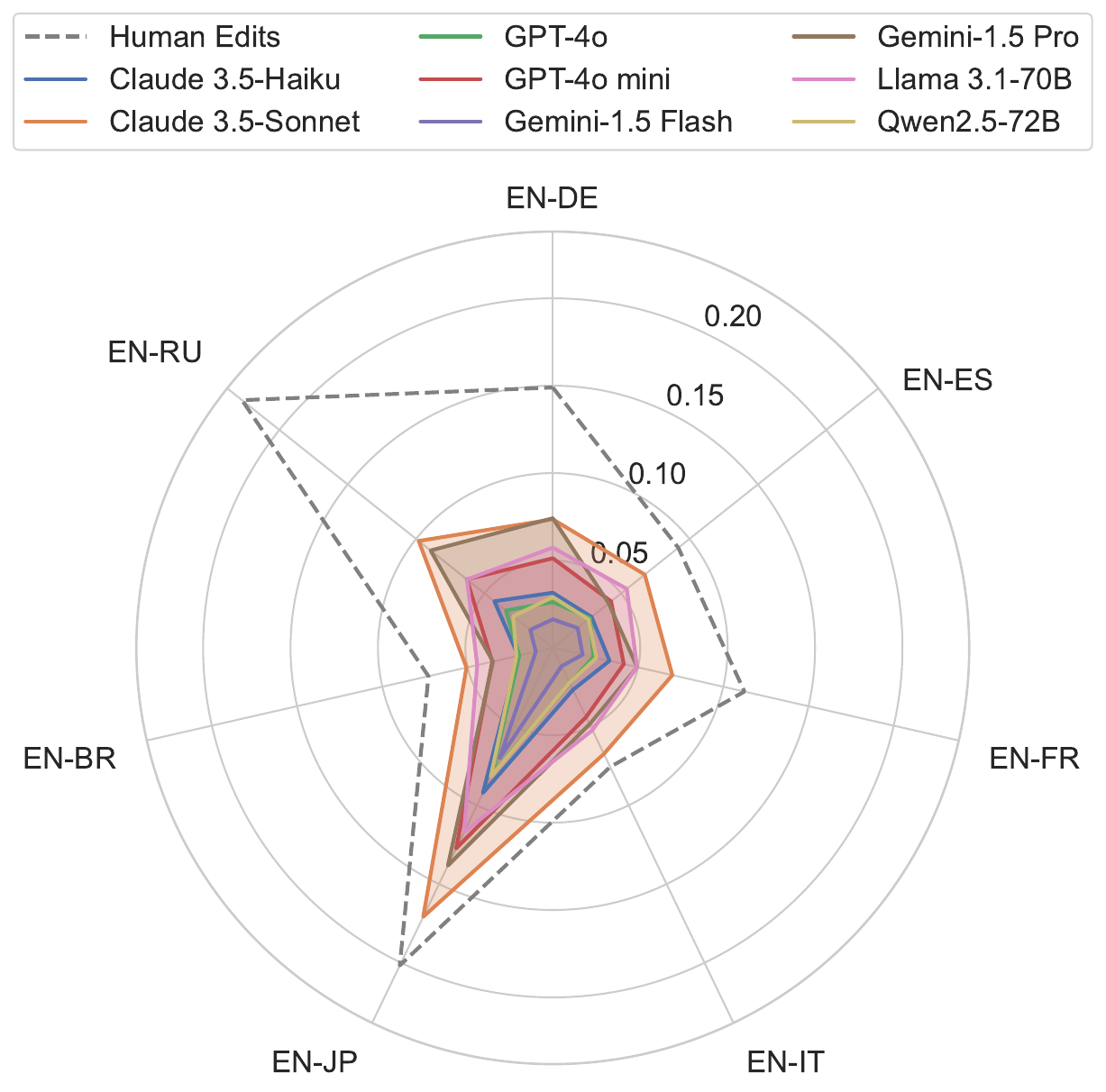}
    \caption{Normalized number of edits made by each model on the NMT output. Note that all models made significantly fewer edits than the human baseline. This indicates that there is still considerable room for improvement}
    \label{fig:edits-radar}
\end{figure}

Figure~\ref{fig:edits-radar} illustrates the correlation between the edits (i.e., deletion, addition, modification) made by the models and those made by human linguists. We observe that \textit{Gemini-1.5 Flash} makes the fewest edits, while \textit{Gemini-1.5 Pro} and \textit{Claude 3.5-Sonnet} show editing behavior more closely aligned with human linguists. Interestingly, even models with the highest number of edits still make fewer changes than the human baseline, highlighting the complexity of this task in \textbf{LangMark}.

\begin{figure}[t!]
    \centering
    \includegraphics[width=0.48\textwidth]{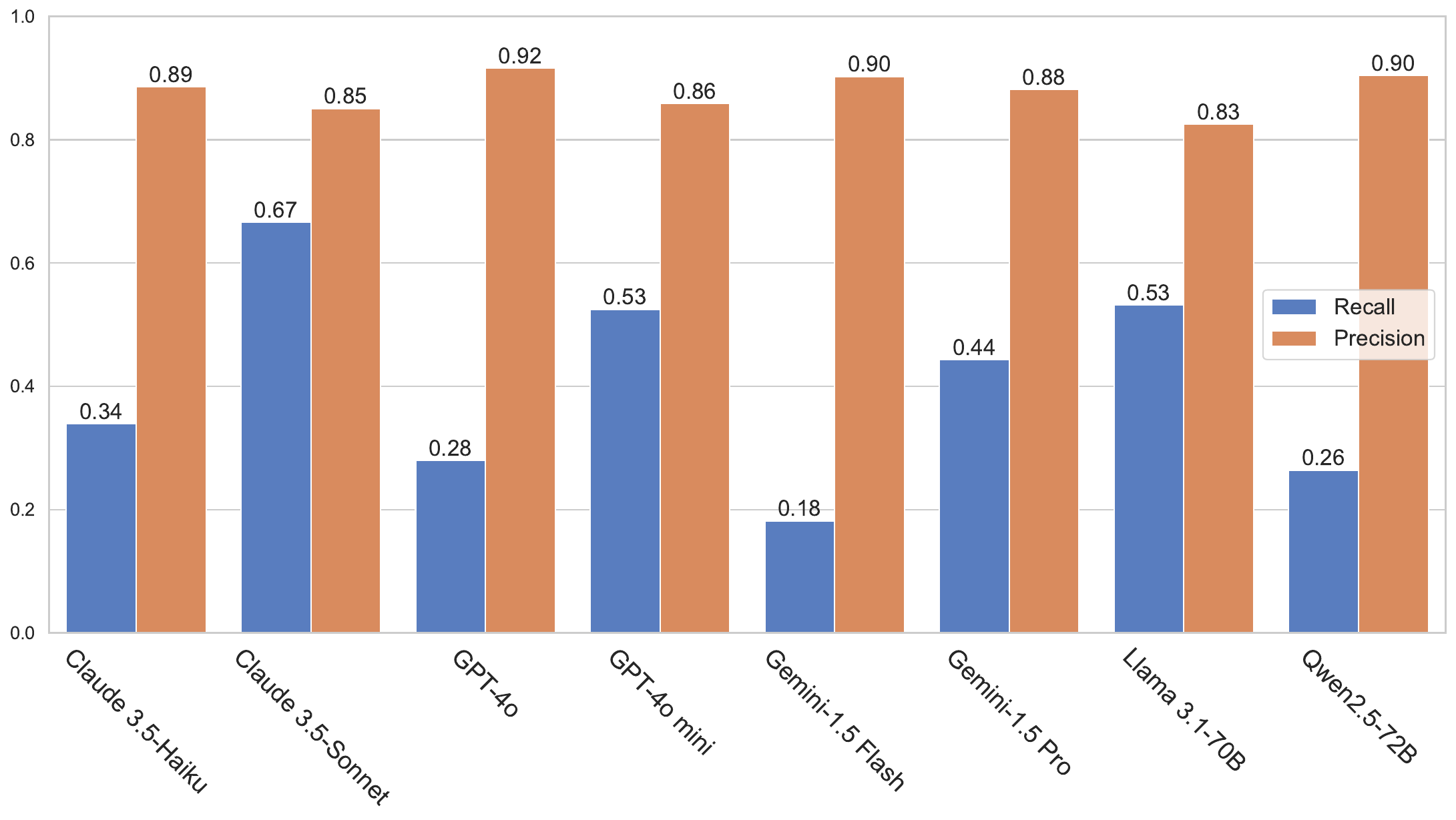}
    \caption{Precision and recall of models when determining that a segment needs to be edited. We see that the models with high recall are not the best performing on machine translation metrics (see Table~\ref{tab:chrf-scores}). Instead, the more ``conservative'' models (low recall, high precision) perform best.}
    \label{fig:edit-detection}
\end{figure}

In the same fashion, Figure~\ref{fig:edit-detection} shows the recall and precision on the triplets that need correction for all models averaged across languages. Note that we do not explicitly prompt the model to classify each triplet. Thus, in this context:

\begin{equation}
\centering
\text{Recall} = \frac{\lvert \{ i \in \mathcal{D} \,|\, {MT}_i \neq {H}_i \land {MT}_i \neq {PE}_i \} \rvert}{\lvert \{ i \in \mathcal{D} \,|\, {MT}_i \neq {H}_i \} \rvert}
\end{equation}

\begin{equation}
\centering
\text{Precision} = \frac{\lvert \{ i \in \mathcal{D} \mid {MT}_i \neq {H}_i \land {MT}_i \neq {PE}_i \}\rvert}{\lvert\{i \in \mathcal{D} \mid {MT}_i \neq {PE}_i \}\rvert}
\end{equation}

Where:

\begin{itemize}[topsep=0pt, partopsep=10pt, itemsep=0pt, parsep=0pt]
    \item \(\mathcal{D}\) is the set of triplets in the dataset.
    \item \(\text{MT}_i\) is the machine translation output for segment \(i\).
    \item \(\text{H}_i\) is the human post-edit (ground truth) for segment \(i\).
    \item \(\text{PE}_i\) is the model post-edit for segment \(i\).
\end{itemize}

Using this formulation, we can quantify both the frequency with which models detect segments that need edits and their accuracy in determining when a segment needs to be edited. Models with higher precision, such as \textit{GPT-4o}, tend to achieve better overall performance on machine translation evaluation metrics despite having lower recall. We refer to these as ``conservative'' models. In contrast, ``aggressive'' models like \textit{Claude 3.5 Sonnet}, perform worse, despite having higher recall.

\begin{figure}[t!]
    \centering
    \includegraphics[width=0.48\textwidth]{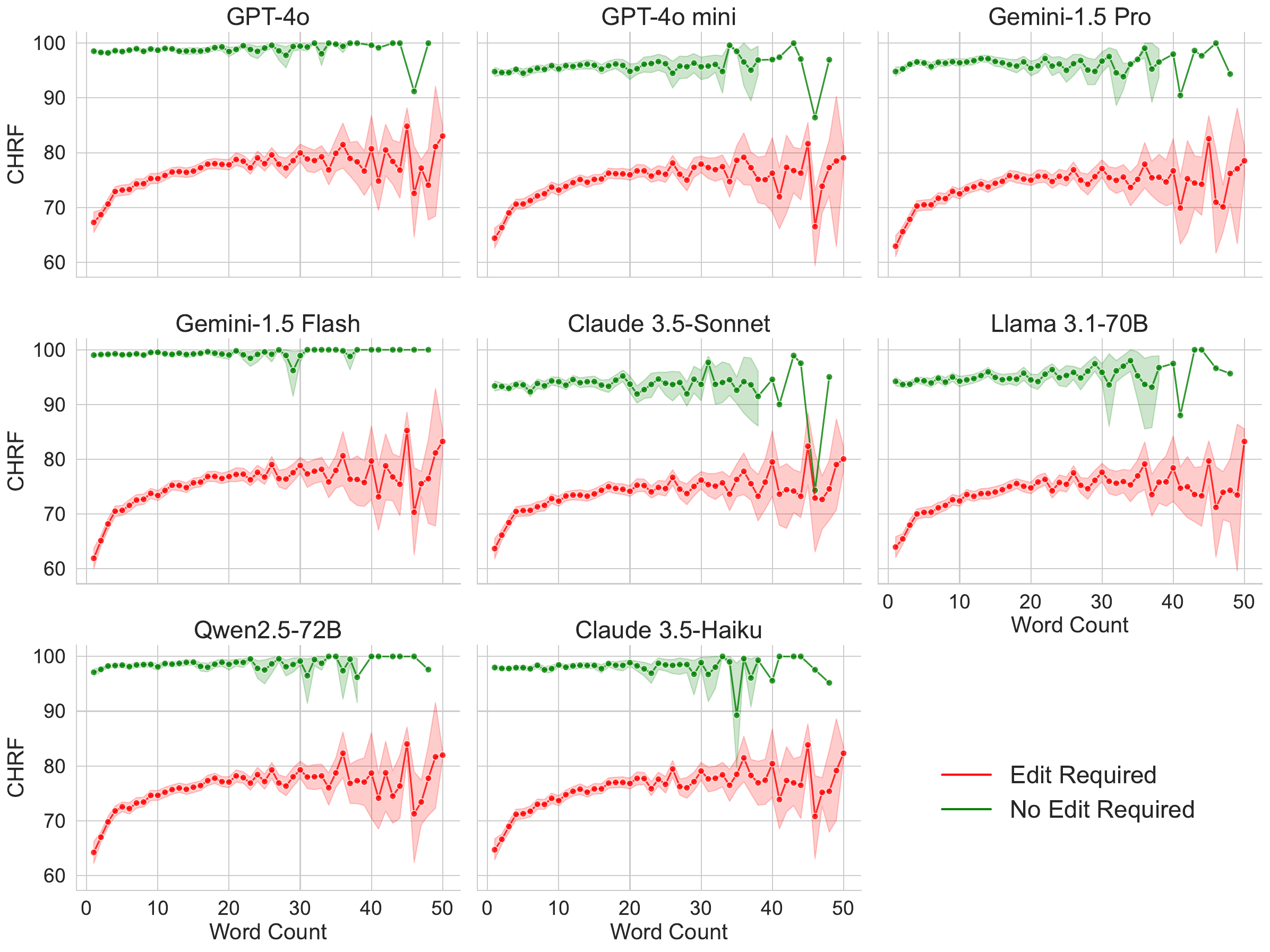}
    \caption{Average performance of each model across segments of varying lengths, separated into those that require edits (\textit{red}) and those that do not (\textit{green}). Models perform substantially worse on shorter segments that need editing, due to limited context. More ``aggressive'' models (\emph{e.g.,} \textit{Claude 3.5 Sonnet}, \textit{GPT-4-mini}) often modify segments that do not require edits. Only segments of up to 50 words are shown for visualization purposes.}
    \label{fig:performance-word}
\end{figure}

Figure~\ref{fig:performance-word} reports the CHRF scores for each model, averaged across all test-set segments and grouped by segment length. For segments requiring no modifications, most models maintain high CHRF scores. However, performance is consistently lower on segments that need correction, hinting at the nuanced nature of the required edits. Editing shorter segments proves especially challenging, likely due to their limited context, which makes it more difficult for APE systems to accurately apply the necessary modifications.

Figures \ref{fig:edit-detection} and \ref{fig:performance-word} show that models with a higher recall often over-detect necessary edits. For instance, \textit{Claude~3.5-Sonnet} identifies more segments that require changes but frequently introduces edits where none are needed, affecting performance. This shows that the task of determining whether a segment requires editing is a key challenge in APE settings, especially when nuanced edits are required. 

\subsection{Towards Better Evaluation Metrics}
These findings suggest that relying solely on machine translation evaluation metrics is insufficient to fully evaluate APE systems. An ideal evaluation metric should consider both the quality of the final output and the number of edits performed, accounting for the balance between unnecessary conservatism and excessive intervention. Although this work does not propose such a metric, we hope that the dataset introduced here fosters further research into the development of comprehensive evaluation frameworks and promotes the design of APE systems that better align with human post-editing strategies.

\section{Conclusions}
This work introduces \textbf{LangMark}, a human-annotated multilingual dataset for automatic post-editing (APE) on neural machine translation (NMT) outputs. The translation is performed {\em from} English to seven languages, and the data is composed of over 200,000 triplets. The dataset and the results presented in this work constitute a valuable benchmark for evaluating APE systems and advancing research in the field.

Our experiments demonstrate that large language models (LLMs) with few-shot prompting can improve translation quality,  outperforming proprietary NMT systems. The fact that most state-of-the-art language models fail to improve on the NMT output that comprises our dataset highlights the strength of \textbf{LangMark} as a benchmark for APE systems. Further, we emphasize that machine translation evaluation metrics, while essential to measure performance, fail to account for the classification part of any APE tasks (i.e., determining whether the NMT output needs to be edited). This highlights the need for metrics that better reflect human editing behavior.

We hope that this dataset and the accompanying analysis provide a foundation for further research and benchmarking of Automatic Post-Editing (APE) systems.


\section*{Limitations}
Although \textbf{LangMark} offers a large-scale, multilingual dataset for automatic post-editing (APE), it also comes with some limitations. First, \textbf{LangMark} is derived from a single domain—marketing content—which may constrain the generalizability of APE models trained on it. The dataset’s linguistic style and error types may not accurately capture challenges in other domains such as medical, legal, or literary texts.

Second, the dataset is unidirectional, covering only translations \emph{from} English \emph{into} seven target languages. This scope excludes the reverse direction (or translations among non-English languages).

Third, our dataset is currently provided in segment-level form rather than as contiguous documents. While this reflects common industry practice (where translators often work on individual segments), it makes direct experimentation with document-level post-editing impossible. We are planning a future release of LangMark that will include full documents, allowing more extensive context for document-level APE experiments.

Fourth, we acknowledge that the usage of a proprietary MT system limits glass-box analysis. While this choice allowed us to create high-quality, challenging APE data, we agree that future work may benefit from including outputs from open-source systems for greater transparency.

Lastly, despite efforts to remove sensitive or personally identifiable information, the original content—drawn from real marketing documents—may still carry domain-specific biases or cultural nuances. Researchers and practitioners should carefully consider these factors when extending or applying \textbf{LangMark} to other use cases or domains.

\section*{Acknowledgments}
We would like to express our gratitude to Smartsheet for providing the resources and data that made this research possible. Their support and collaboration were instrumental in the development of the multilingual automatic post-editing dataset presented in this paper. This work would not have been possible without their commitment to advancing research in the field of natural language processing and machine translation.

\bibliography{mtsummit25}

\newpage
\appendix
\onecolumn
\section{Appendix}
\label{sec:appendix}

\subsection{Linguist Compensation}
\label{sec:comp}
In terms of our freelance supplier pool, we prioritize fair compensation for our linguists based on the complexity of their tasks and prevailing market rates. We ensure that our pay rates reflect the market value for each language combination and required skill set, guaranteeing equitable remuneration for all services provided.
 
Beyond fair pay, we are dedicated to supporting local rural communities in India and Africa through our impactful sourcing program. This initiative creates valuable opportunities for individuals in marginalized communities who might not otherwise have access to such work. Currently, we are running three successful programs in collaboration with companies in these regions.
 
Additionally, we place great emphasis on engaging with our linguist community. We regularly conduct surveys to gather feedback and continuously refine our work practices, ensuring we meet the needs and expectations of our talented linguists. 

\subsection{Instructions for Linguists}
\label{sec:instructions}
This section provides an overview of the instructions given to linguists assigned for post-editing. After informing the linguists of the general task they will be performing \emph{i.e., Post-editing for a given project}, linguist are instructed as follows:

You will work online in the usual TMS with little changes in the overall workflow process.
TM is still leveraged for matches $\geq 75\%$ and MT will only be leveraged for all No Matches.
You will review Fuzzy TM matches and post-edit MT segments to meet the agreed quality level.

\paragraph{Quality Expectations} MT is from a General Neural MT System, which means you must pay special attention to terminology, which may not be compliant with the client/domain specifications.
Post-edited translations must comply with all current reference material, such as style guides, glossaries, DNT lists, UI references or other project-specific instructions.
Full post-editing should produce semantically accurate translations that consistently use correct and 
approved terminology and are free from grammatical errors.
The translation should have the appropriate tone and style for the given content and read as if written in the 
target language.

\paragraph{Best Practice Post-Editing Steps} Here are some steps to guarantee quality in the post editing process:
\begin{itemize}
    \item \textbf{READ:} Compare the source and the machine translation suggestion. Decide quickly which parts of the MT can be used.
    \item \textbf{EDIT:} Make changes to MT where necessary, using as much of the MT output as possible. Use the good “bits/sections”, move them around, correct word forms, change parts of speech, and use them as inspiration for your translation.
    \item \textbf{QA:} Look up key terms in your reference material as usual to ensure terminology consistency and compliance with TM, glossary, DNT list, UI references. Perform standard QA checks, and ensure spelling and punctuation are as required for regular translation.
\end{itemize}

\paragraph{Typical Errors in Neural MT}
Being aware of typical errors helps good post-editing. Depending on the language pair, there are typical errors to fix:

\begin{itemize}
    \item Capitalization, such as missing or inconsistent capitalization in UI options or product names
    \item Word order in MT output may follow the source and needs to be rearranged per target language rules
    \item Spacing \& Punctuation, may be following the source or not be compliant with target language rules
    \item Errors in word form agreement, such as gender, number or case mismatch
    \item Additions (content or words added in the MT that are not in the source)
    \item Measurements, dates and other numerals may need to be adapted as required by client guidelines
    \item Omissions (content in the source that is missing in the MT)
    \item Wrong or inconsistent terminology, or terminology that is correct but not compliant with client specifications
    \item New words that the MT engine has not encountered before may be left untranslated or mistranslated
    \item Tags or placeholders may be missing or incorrectly placed in the MT, or text within them has been translated
\end{itemize}

\subsection{Post Editing Examples}
\label{sec:pe_examples}
Below we present several examples of edits done by \textbf{GPT-4o} to the NMT. Each box contains the source, the NMT (Pre-Translation), and the improved Post-Edit along with a short explanation.

\begingroup
\footnotesize
\begin{tcbraster}[raster columns=2, raster equal height=rows,
                  raster column skip=3mm, raster row skip=3mm]
\begin{tcolorbox}[nobeforeafter, title=German (de-DE),
    boxsep=2pt,
    arc=2pt,
    left=3pt,
    right=3pt,
    colback=gray!5,
    colframe=gray!75!black]
\textbf{Source:}\\
\emph{Pro Desk Session: Smartsheet Advance Capabilities}\\[3pt]
\textbf{Pre-Translation:}\\
Pro Desk Session: Smartsheet Advance Fähigkeiten\\[3pt]
\textbf{LLM Post-Edit:}\\
Pro Desk Session: Smartsheet Advance Funktionen\\[3pt]
\textbf{Explanation:}\\
Changed \emph{Fähigkeiten} (abilities) to \emph{Funktionen} (features).
\end{tcolorbox}
\begin{tcolorbox}[nobeforeafter,  title=Japanese (ja-JP),
    boxsep=2pt,
    arc=2pt,
    left=3pt,
    right=3pt,
    colback=gray!5,
    colframe=gray!75!black]
\textbf{Source:}\\
\emph{\{\{filterName\}\} is private.}\\[3pt]
\textbf{Pre-Translation:}\\
\begin{CJK}{UTF8}{min}
  \% は非公開です。
\end{CJK}\\[3pt]
\textbf{LLM Post-Edit:}\\
\begin{CJK}{UTF8}{min}
  \{\{filterName\}\} は非公開です。
\end{CJK}\\[3pt]
\textbf{Explanation:}\\
Replaced the incorrect placeholder ``\%'' with \{\{filterName\}\}.
\end{tcolorbox}
\begin{tcolorbox}[nobeforeafter, title=Portuguese (pt-BR),
    boxsep=2pt,
    arc=2pt,
    left=3pt,
    right=3pt,
    colback=gray!5,
    colframe=gray!75!black]
\textbf{Source:}\\
\emph{If you’re already logged in, you might see a new tab opening instead of a new window. That’s normal.}\\[3pt]
\textbf{Pre-Translation:}\\
Se você já estiver conectado, poderá ver uma nova guia em vez de uma nova janela. Isso é normal.\\[3pt]
\textbf{LLM Post-Edit:}\\
Se você já estiver conectado, poderá ver uma nova guia abrindo em vez de uma nova janela. Isso é normal.\\[3pt]
\textbf{Explanation:}\\
Rephrased for a more natural tone in Brazilian Portuguese.
\end{tcolorbox}
\begin{tcolorbox}[nobeforeafter, title=Italian (it-IT),
    boxsep=2pt,
    arc=2pt,
    left=3pt,
    right=3pt,
    colback=gray!5,
    colframe=gray!75!black]
\textbf{Source:}\\
\emph{Create reports and dashboards in a workspace or start from scratch.}\\[3pt]
\textbf{Pre-Translation:}\\
Crea report e pannelli di controllo in un workspace o inizia da zero.\\[3pt]
\textbf{LLM Post-Edit:}\\
Crea report e dashboard in un workspace o inizia da zero.\\[3pt]
\textbf{Explanation:}\\
Changed \emph{pannelli di controllo} to \emph{dashboard}.
\end{tcolorbox}
\begin{tcolorbox}[
    title=Spanish (es-419),
    boxsep=2pt,
    arc=2pt,
    left=3pt,
    right=3pt,
    colback=gray!5,
    colframe=gray!75!black
]
\textbf{Source:}\\
\emph{Unpin}\\[3pt]
\textbf{Pre-Translation:}\\
Desmarcar\\[3pt]
\textbf{LLM Post-Edit:}\\
Desanclar\\[3pt]
\textbf{Explanation:}\\
Correct verb for ``unpin'' in Spanish is \emph{desanclar}.
\end{tcolorbox}
\begin{tcolorbox}[nobeforeafter, title=French (fr-FR),
    boxsep=2pt,
    arc=2pt,
    left=3pt,
    right=3pt,
    colback=gray!5,
    colframe=gray!75!black]
\textbf{Source:}\\
\emph{Excel Calendar and Checklist Templates}\\[3pt]
\textbf{Pre-Translation:}\\
Modèles d’agenda et de liste de contrôle Excel\\[3pt]
\textbf{LLM Post-Edit:}\\
Modèles de calendrier et de liste de contrôle Excel\\[3pt]
\textbf{Explanation:}\\
Switched from \emph{agenda} to \emph{calendrier} to better reflect ``calendar.''
\end{tcolorbox}

\end{tcbraster}
\endgroup

\noindent
Across different examples, we can see various types of fixes, including correction of placeholders, terminology, style, and usage in diverse contexts.
\newpage
\subsection{Zero-Shot Results}
\label{sec:zero-shot}

\begin{table}[!h]
\centering  
\centering
\caption{Zero-shot CHRF scores for different models and languages when performing APE on the test set. Scores are compared across models, with the proprietary MT serving as the baseline.}
\resizebox{0.7\textwidth}{!}{ 
\begin{tabular}{lccccccc}
\toprule
 & & \multicolumn{6}{c}{\textbf{Languages}} \\
\cmidrule(lr){2-8}
\textbf{Model} & \textbf{EN-RU} & \textbf{EN-PT} & \textbf{EN-JP} & \textbf{EN-IT} & \textbf{EN-FR} & \textbf{EN-ES} & \textbf{EN-DE} \\
\midrule
\textbf{Baseline} & 68.90 & \textbf{89.44} & 70.22 & \textbf{89.58} & 81.96 & 86.07 & \textbf{81.29} \\[0.2em]
\midrule
\textbf{Gemini-1.5 Flash} & 68.80 & 88.97 & 71.59 & 88.95 & 82.26 & 86.14 & 80.85 \\
\textbf{Gemini-1.5 Pro} & 65.95 & 86.65 & 68.01 & 84.42 & 79.74 & 84.45 & 77.67 \\
\textbf{Claude 3.5-Sonnet} & 67.83 & 87.68 & 68.00 & 86.78 & 80.73 & 83.43 & 79.18 \\
\textbf{Claude 3.5-Haiku} & 68.62 & 88.86 & 71.90 & 88.99 & 82.24 & 86.01 & 80.57 \\
\textbf{GPT-4o mini} & 67.78 & 87.84 & 69.73 & 87.99 & 81.40 & 84.91 & 80.10 \\
\textbf{GPT-4o} & \textbf{68.99} & 89.21 & \textbf{73.46} & 89.29 & \textbf{82.24} & \textbf{86.34} & 81.06 \\
\midrule
\multicolumn{8}{c}{\textbf{Open Source}} \\
\midrule
\textbf{Llama 3.1-70B} & 66.84 & 85.41 & 68.80 & 85.30 & 79.88 & 81.54 & 77.07 \\
\textbf{Qwen2.5-72B} & 68.62 & 89.21 & 72.86 & 89.23 & 82.27 & 86.07 & 81.08 \\
\bottomrule
\end{tabular}
}
\label{tab:chrf-scores-zero_shot}
\end{table}

\begin{table}[!h]
\centering  
\caption{Zero-shot TER$\downarrow$ \citep{snover-etal-2006-study} scores for different models and languages when performing APE on the test set. Scores are compared across models, with the proprietary MT serving as the baseline.}
\scriptsize
\resizebox{0.7\textwidth}{!}{
\begin{tabular}{lccccccc}
\toprule
 & & \multicolumn{6}{c}{\textbf{Languages}} \\
\cmidrule(lr){2-8}
\textbf{Model} & \textbf{EN-RU} & \textbf{EN-PT} & \textbf{EN-JP} & \textbf{EN-IT} & \textbf{EN-FR} & \textbf{EN-ES} & \textbf{EN-DE} \\
\midrule
\textbf{Baseline} & 45.40 & 14.27 & 74.15 & \textbf{14.61} & 26.67 & 19.28 & \textbf{31.26} \\[0.2em]
\midrule
\textbf{Gemini-1.5 Flash} & 45.71 & 14.67 & 72.87 & 15.40 & 25.60 & 19.28 & 31.61 \\
\textbf{Gemini-1.5 Pro} & 49.51 & 17.65 & 74.52 & 20.94 & 28.76 & 21.42 & 35.77 \\
\textbf{Claude 3.5-Sonnet} & 47.16 & 16.18 & 79.14 & 18.24 & 27.70 & 22.75 & 33.74 \\
\textbf{Claude 3.5-Haiku} & 45.70 & 14.66 & 74.75 & 15.28 & \textbf{25.56} & 19.41 & 31.76 \\
\textbf{GPT-4o mini} & 46.66 & 15.63 & 76.08 & 16.52 & 26.52 & 20.58 & 32.47 \\
\textbf{GPT-4o} & \textbf{45.35} & 14.67 & \textbf{71.75} & 14.96 & 25.87 & \textbf{19.04} & 31.30 \\
\midrule
\multicolumn{8}{c}{\textbf{Open Source}} \\
\midrule
\textbf{Llama 3.1-70B} & 47.77 & 18.67 & 76.20 & 19.59 & 28.83 & 27.85 & 41.08 \\
\textbf{Qwen2.5-72B} & 45.69 & \textbf{14.22} & 71.25 & 15.00 & 25.66 & 19.34 & 31.30 \\
\bottomrule
\end{tabular}
}
\label{tab:ter-scores-zero_shot}
\end{table}

\begin{table}[!t]
\centering  
\caption{Zero-shot BLEU \citep{papineni2002bleu} scores for different models and languages when performing APE on the test set. Scores are compared across models, with the proprietary MT serving as the baseline.}
\resizebox{0.7\textwidth}{!}{
\begin{tabular}{lccccccc}
\toprule
 & & \multicolumn{6}{c}{\textbf{Languages}} \\
\cmidrule(lr){2-8}
\textbf{Model} & \textbf{EN-RU} & \textbf{EN-PT} & \textbf{EN-JP} & \textbf{EN-IT} & \textbf{EN-FR} & \textbf{EN-ES} & \textbf{EN-DE} \\
\midrule
\textbf{Baseline} & \textbf{49.13} & \textbf{80.16} & 14.28 & \textbf{79.93} & \textbf{64.91} & 73.75 & \textbf{64.13} \\[0.2em]
\midrule
\textbf{Gemini-1.5 Flash} & 48.90 & 79.51 & 33.61 & 79.09 & 66.56 & 74.28 & 63.61 \\
\textbf{Gemini-1.5 Pro} & 44.31 & 75.31 & 32.80 & 71.28 & 62.68 & 71.44 & 58.34 \\
\textbf{Claude 3.5-Sonnet} & 47.44 & 77.12 & 30.93 & 75.34 & 64.44 & 69.76 & 60.82 \\
\textbf{Claude 3.5-Haiku} & 48.63 & 79.37 & 33.38 & 79.13 & 66.73 & 74.06 & 63.20 \\
\textbf{GPT-4o mini} & 47.62 & 77.69 & 27.51 & 77.47 & 65.37 & 72.30 & 62.40 \\
\textbf{GPT-4o} & 48.99 & 79.58 & \textbf{34.95} & 79.51 & 66.02 & \textbf{74.49} & 63.82 \\
\midrule
\multicolumn{8}{c}{\textbf{Open Source}} \\
\midrule
\textbf{Llama 3.1-70B} & 46.03 & 73.87 & 32.31 & 73.17 & 63.03 & 65.58 & 54.83 \\
\textbf{Qwen2.5-72B} & 48.45 & 79.79 & 34.24 & 79.46 & 66.62 & 74.20 & 63.90 \\
\bottomrule
\end{tabular}
}
\label{tab:bleu-scores-zero_shot}
\end{table}
\newpage
\subsection{Additional Metrics}
\label{sec:more-metrics}

\begin{table}[!h]
\centering  
\caption{TER$\downarrow$ scores \citep{snover-etal-2006-study} for different models and languages when performing APE on the test set. Scores are compared across models, with the proprietary MT serving as the baseline. Lower is better.}
\scriptsize
\resizebox{0.7\textwidth}{!}{
\begin{tabular}{lccccccc}
\toprule
 & & \multicolumn{6}{c}{\textbf{Languages}} \\
\cmidrule(lr){2-8}
\textbf{Model} & \textbf{EN-RU} & \textbf{EN-PT} & \textbf{EN-JP} & \textbf{EN-IT} & \textbf{EN-FR} & \textbf{EN-ES} & \textbf{EN-DE} \\
\midrule
\textbf{Baseline} & 45.40 & \textbf{14.27} & 74.15 & 14.61 & 26.67 & 19.28 & 31.26\\[0.2em]
\midrule
\textbf{Gemini-1.5 Flash} & 45.62 & 14.42 & 71.59 & 14.81 & 25.83 & 19.14 & 31.43 \\
\textbf{Gemini-1.5 Pro} & 47.53 & 16.37 & 70.84 & 19.52 & 27.95 & 20.76 & 35.60 \\
\textbf{Claude 3.5-Sonnet} & 46.56 & 17.82 & 75.66 & 20.57 & 28.34 & 23.67 & 34.90 \\
\textbf{Claude 3.5-Haiku} & 45.60 & 14.72 & 72.12 & 15.59 & 25.71 & 19.51 & 31.78 \\
\textbf{GPT-4o mini} & 46.17 & 16.08 & 74.68 & 17.27 & 26.54 & 20.56 & 32.74 \\
\textbf{GPT-4o} & 44.49 & 14.41 & 69.01 & \textbf{14.25} & \textbf{25.30} & \textbf{18.64} & \textbf{30.91} \\
\midrule
\multicolumn{8}{c}{\textbf{Open Source}} \\
\midrule
\textbf{Llama 3.1-70B} & 45.12 & 17.44 & 73.94 & 18.39 & 27.80 & 22.26 & 33.80 \\
\textbf{Qwen2.5-72B} & \textbf{43.91} & 14.45 & \textbf{68.75} & 15.23 & 25.71 & 18.95 & 30.95 \\
\bottomrule
\end{tabular}
}
\label{tab:ter-scores}
\end{table}

\begin{table}[!h]
\centering  
\caption{BLEU \citep{papineni2002bleu} scores for different models and languages when performing APE on the test set. Scores are compared across models, with the proprietary MT serving as the baseline.}
\scriptsize
\resizebox{0.7\textwidth}{!}{
\begin{tabular}{lccccccc}
\toprule
 & & \multicolumn{6}{c}{\textbf{Languages}} \\
\cmidrule(lr){2-8}
\textbf{Model} & \textbf{EN-RU} & \textbf{EN-PT} & \textbf{EN-JP} & \textbf{EN-IT} & \textbf{EN-FR} & \textbf{EN-ES} & \textbf{EN-DE} \\
\midrule
\textbf{Baseline} & 49.13 & \textbf{80.16} & 14.28 & 79.93 & 64.91 & 73.75 & 64.13 \\[0.2em]
\midrule
\textbf{Gemini-1.5 Flash} & 48.69 & 79.80 & 34.17 & 79.59 & 66.50 & 74.37 & 63.71 \\
\textbf{Gemini-1.5 Pro} & 46.35 & 77.04 & 36.27 & 73.23 & 63.74 & 72.47 & 58.16 \\
\textbf{Claude 3.5-Sonnet} & 47.53 & 74.83 & 33.94 & 71.92 & 63.61 & 68.20 & 59.08 \\
\textbf{Claude 3.5-Haiku} & 48.58 & 79.17 & 35.72 & 78.72 & 66.61 & 74.11 & 63.10 \\
\textbf{GPT-4o mini} & 47.92 & 77.30 & 27.81 & 76.17 & 65.21 & 72.27 & 61.89 \\
\textbf{GPT-4o} & 49.79 & 79.86 & \textbf{37.96} & \textbf{80.12} & \textbf{66.91} & \textbf{74.84} & \textbf{64.20} \\
\midrule
\multicolumn{8}{c}{\textbf{Open Source}} \\
\midrule
\textbf{Llama 3.1-70B} & 49.28 & 75.76 & 33.01 & 74.97 & 64.22 & 70.27 & 60.70 \\
\textbf{Qwen2.5-72B} & \textbf{50.31} & 79.59 & 37.43 & 79.16 & 66.60 & 74.79 & 64.01 \\
\bottomrule
\end{tabular}
}
\label{tab:bleu-scores}
\end{table}

\end{document}